\documentclass[11pt]{article}
\usepackage[margin=1in]{geometry}
\usepackage{amsmath,amssymb,graphicx,booktabs,longtable,array}
\usepackage[hidelinks]{hyperref}
\title{A JiT-Based Denoising Defense Method Against Adversarial Examples for Radio Signal Automatic Modulation Classification}
\author{Shiwei Liu \and Linyuan Wang\thanks{Corresponding author.} \and Nuolin Sun \and Libin Hou \and Bin Yan}
\date{}
\begin{document}
\maketitle
\begin{center}\small Henan Key Laboratory of Imaging and Intelligent Processing,
Information Engineering University, Zhengzhou 450000, China\end{center}

\begin{abstract}

Automatic modulation classification (AMC) of radio signals is a core technology for spectrum monitoring and electromagnetic situational awareness. Deep neural networks (DNNs) play an important role in this field owing to their powerful feature extraction capabilities and end-to-end learning paradigm. However, their vulnerability to adversarial attacks poses serious security challenges to practical deployment. Adversarial training, a common defense, has limitations including poor generalization across attack types and high training costs. Inspired by Just image Transformers (JiT), a diffusion generative model that directly predicts clean images, this paper proposes a JiT-based denoising defense against adversarial examples for radio signal modulation classification. During training, a signal-specific 1D Transformer denoising network is optimized via flow matching to directly predict clean signal samples. It thereby learns a mapping that projects arbitrary examples back onto the low-dimensional manifold of clean signal examples. During inference, arbitrary examples are fed into the trained network and directly projected onto this manifold. This enables denoising defense without prior knowledge of attacks. Experiments on the RML2018.01A dataset were conducted without attack information, using a single empirically selected, fixed denoising start time. Compared with PGD-based adversarial training, the proposed method achieves a relative improvement of at least 20.58\% in robust accuracy under PGD attacks. It also achieves a relative improvement of 10.20\% in classification accuracy on clean examples. When evaluated against attack types unseen during adversarial training, the proposed method provides stronger defense than adversarial training. In particular, at higher attack strengths, it achieves a relative improvement of at least 39.72\% in robust accuracy.

\end{abstract}


\section{Introduction}

Automatic modulation classification (AMC) of radio signals is a core technology for spectrum monitoring and electromagnetic situational awareness~\cite{ref1,ref2}. As intelligent information processing technologies become more widely applied in radio signal sensing, AMC plays an increasingly important role. In recent years, deep neural networks (DNNs) have achieved recognition accuracy on AMC tasks far exceeding that of traditional methods because of their powerful end-to-end feature extraction capabilities~\cite{ref3,ref4,ref5}. They have therefore become an important approach for intelligent radio signal processing. From the perspective of intelligent-system security, adversarial examples pose a serious challenge to DNN robustness. By adding carefully designed and imperceptible small perturbations to input signals, an attacker can cause a classifier to output an incorrect modulation class with high confidence~\cite{ref6}. This vulnerability creates serious security threats in open radio spectrum environments. Attackers can interfere with the modulation classification of legitimate receivers by imposing adversarial perturbations on target signals. This can disrupt the normal operation of communication links or enable signal spoofing~\cite{ref7}. Therefore, research on defenses against adversarial examples for radio signal modulation classification is important for effectively countering the security threats caused by adversarial perturbations and ensuring communication security in complex electromagnetic environments.

In recent years, research on defenses against adversarial examples in signals has attracted increasing attention. For signal preprocessing defenses, Zhang et al.~\cite{ref8} proposed the Homomorphic Filtering Adversarial Defense (HFAD) method. It suppresses high-frequency perturbation components in signals and combines data augmentation to improve model robustness. Dong et al.~\cite{ref9} proposed using feature contribution analysis to identify sampling positions that have significant negative effects on classification decisions. The identified positions are then used for input-data purification. Hong et al.~\cite{ref10} proposed a reconstruction-based defense framework for automatic modulation recognition. The framework reconstructs input signals to recover the semantic features and physical consistency of clean examples, thereby improving recognition performance under adversarial attacks. Most such methods locally modify examples in the input space. Their explicit modeling of the overall distribution of clean signals is relatively limited. For research on data manifolds and latent spaces, Zhang et al.~\cite{ref11} proposed the Disentangled Information Bottleneck Adversarial Defense (DIBAD) method. Under an information bottleneck constraint, this method uses the Hilbert-Schmidt independence criterion to separate task-related features from perturbation-related redundant features. It thereby suppresses the propagation of adversarial perturbations at the latent representation level. Zhang et al.~\cite{ref12} proposed the Collaborative VQMAE method. It uses a vector-quantized masked autoencoder to represent and recover perturbed signals in the latent representation space, thereby improving adversarial robustness. Such methods generally rely on latent-space constraints, feature disentanglement, or encoding and decoding processes. Under strong attacks, their defense performance may decline because of incomplete feature disentanglement and error propagation through the reconstruction and quantization stages.

A comprehensive analysis of existing studies shows that adversarial training~\cite{ref13} remains a well-validated and widely used approach for signal adversarial defense. It is also an important baseline for performance comparisons in most related studies~\cite{ref14}. Bao et al.~\cite{ref15} proposed the Optimizing Adversarial Training via Genetic Algorithm (OATGA) method. This method introduces a genetic algorithm into the adversarial training process and improves the search for and optimization of adversarial examples during training, thereby enhancing the robustness of AMC models. Zhan et al.~\cite{ref16} used an attention-guided mechanism to suppress low-importance perturbation features. When combined with adversarial training, this mechanism achieved improved defense performance. However, as attack techniques continue to evolve, the limitations of adversarial training in improving model robustness have become increasingly evident. Its adaptability is limited. Most adversarial training methods provide effective defense against specific attack types encountered during training, but generalize poorly to attack types unseen by the network. Its training cost is also high. Generating adversarial examples for training is time-consuming and computationally expensive.

The generative diffusion model JiT~\cite{ref17} introduces the direct prediction of clean examples by diffusion models. This enables networks to effectively learn the low-dimensional manifold structure on which clean data lie in high-dimensional pixel spaces~\cite{ref18,ref19,ref20,ref21}. This study considers adversarial examples to be noisy examples that deviate from the manifold of clean examples. If a denoising network can be trained to learn an effective mapping that projects arbitrary examples back onto the low-dimensional manifold of clean examples, adversarial examples can be directly projected onto this manifold to achieve efficient denoising defense. Based on this idea, this paper proposes an effective strategy for adversarial-example defense. As shown in Fig. 1, the denoising generation process of the JiT model is adopted. The network is trained through flow matching~\cite{ref22,ref23} to directly learn and predict clean signals. It thereby learns a mapping that projects arbitrary examples back onto the low-dimensional manifold of clean examples. Based on the learned mapping, an arbitrary example is directly used as the starting point for denoising without injecting additional noise. The example is then projected onto the low-dimensional manifold on which clean examples lie. This achieves direct denoising defense. The main contributions of this work are as follows:

\begin{enumerate}

\item We propose an effective denoising defense method against adversarial examples, inspired by the JiT denoising process. By predicting clean examples, the network learns a mapping that projects arbitrary examples back onto the low-dimensional manifold of clean examples. This enables denoising by projecting arbitrary examples onto the clean-example manifold, with the learned distribution of clean examples serving as the prior.

\item We construct a one-dimensional Transformer denoising network, based on the characteristics of I/Q data in radio signals. The network is trained through flow matching to learn a mapping that projects arbitrary examples back onto the low-dimensional manifold of clean examples. Adversarial examples can be directly used as denoising starting points without injecting additional noise, enabling projection-based denoising.

\item On the RML2018.01A dataset~\cite{ref24}, the proposed ``direct denoising'' defense strategy empirically selects a fixed denoising start time without access to attack information. Compared with adversarial training based on PGD attacks, it achieves a relative improvement of at least 20.58\% in robust accuracy under PGD attacks. It also achieves a relative improvement of 10.20\% in clean-example recognition accuracy. For attack types unseen during adversarial training, the proposed method exhibits more pronounced defense advantages. In particular, at higher attack strengths, it achieves a relative improvement of at least 39.72\% in robust accuracy.

\end{enumerate}

\section{Methods}

Inspired by the direct prediction of clean examples during denoising in the diffusion generative model JiT, we construct a 1D Transformer denoising network during training. The network is trained using flow matching to directly learn and predict clean signal examples. Through this training, it learns a mapping that projects arbitrary examples back onto the low-dimensional manifold of clean examples. During inference, we consider adversarial examples to be noisy examples that deviate from the manifold of clean signals. Using adversarial examples directly as denoising starting points, the trained denoising network projects them onto the low-dimensional clean-signal manifold, enabling direct and efficient denoising defense.

\subsection{Construction of the Signal Denoising Network}

\begin{figure}[!htbp]
\centering
\includegraphics[width=.68\linewidth,height=.31\textheight,keepaspectratio]{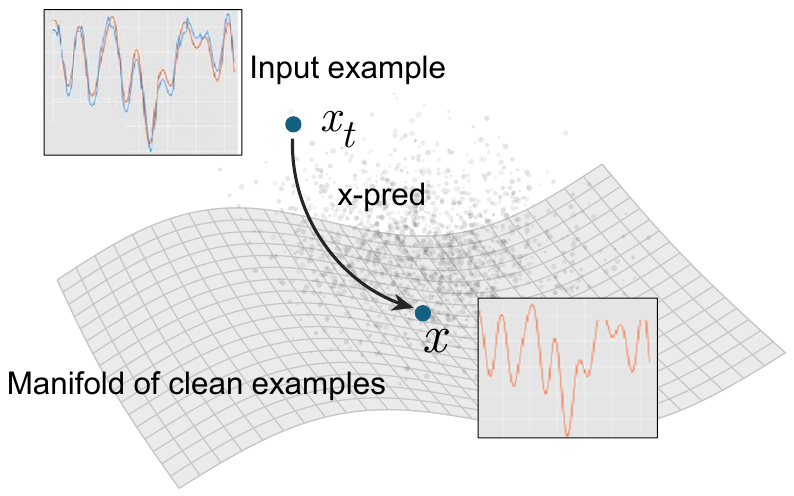}
\caption{Schematic illustration of the JiT denoising defense method.}\label{fig:1}
\end{figure}

The JiT study~\cite{ref17} indicates that clean examples are not uniformly distributed in their high-dimensional ambient space, they are concentrated on a low-dimensional manifold embedded in that space. This paper assumes that radio modulation signal data in I/Q form exhibit low-dimensional manifold embedding characteristics similar to those of natural image data. If a network model can be trained to learn a mapping that projects arbitrary examples back onto the low-dimensional manifold of clean examples, this mapping enables denoising. Any example outside the manifold can then be projected back onto the clean-example manifold through this mapping. This projection-based defense concept is illustrated in Fig.~\ref{fig:1}.

We design a denoising network specifically tailored to the characteristics of signal data. Its detailed architecture is shown in Fig. 2. Inspired by the bottleneck embedding design in JiT~\cite{ref17}, we develop a two-layer bottleneck convolutional embedding module (BottleneckPatchEmbed1D). This design helps the network better learn the mapping that projects arbitrary examples back onto the low-dimensional manifold of clean examples. For input I/Q signals with a shape of 1024 $\times$ 2, a one-dimensional convolution with a patch length of 16 is first used for patch embedding. This represents each signal as 64 local segments and maps the feature dimension of each segment from 2 to 32. A subsequent 1 $\times$ 1 convolution further increases the embedding dimension to 256, yielding a 64 $\times$ 256 feature sequence. The data then pass through a six-layer 1D Transformer feature extraction network with a hidden dimension of 256. Eight-head self-attention is used to model correlations across segments. Finally, the 1024-dimensional signal data are reconstructed through a linear mapping

In this study, the denoising network is trained using the same flow matching method as the generative model JiT~\cite{ref17}. By predicting clean examples, the network learns a mapping that projects arbitrary examples back onto the low-dimensional manifold of clean examples. An arbitrary example $x_t$ outside the low-dimensional manifold is represented as a linear interpolation between a clean example $x$ and noise $\varepsilon$, $t\in[0,1]$.

\begin{equation}\label{eq:1}
x_t=tx+(1-t)\varepsilon,\qquad t\in[0,1]
\end{equation}

When $t=1$, $x_t$ lies on the low-dimensional manifold in which clean examples are embedded. When $t<1$, $x_t$ gradually deviates from the low-dimensional manifold. During the projection of examples onto the low-dimensional manifold of clean examples, the change in $x_t$ over time is defined as the flow velocity, as follows:

\begin{equation}\label{eq:2}
v=\frac{dx_t}{dt}=x-\varepsilon=\frac{x-x_t}{1-t}
\end{equation}

Accordingly, the relationship between the clean signal and the target flow velocity is established. When the network predicts the clean example $x_\theta$, the corresponding flow velocity is expressed as follows:

\begin{equation}\label{eq:3}
v_\theta(x_t,t)=\frac{x_\theta-x_t}{1-t}
\end{equation}

This establishes the following mean squared error loss function for the flow velocity used to train the denoising network:

\begin{equation}\label{eq:4}
L=\mathbb{E}_{t,x,\varepsilon}\left\lVert v_\theta(x_t,t)-v\right\rVert^2
\end{equation}

Learning and predicting clean examples can be achieved by minimizing the flow-velocity loss function $L$. Here, $v_\theta(x_t,t)$ denotes the flow velocity predicted by the network, $v$ denotes the target flow velocity, and $\theta$ denotes the network parameters. The detailed training procedure is presented in the algorithmic workflow in Table 1.

\begin{figure}[!htbp]
\centering
\includegraphics[height=.50\textheight,keepaspectratio]{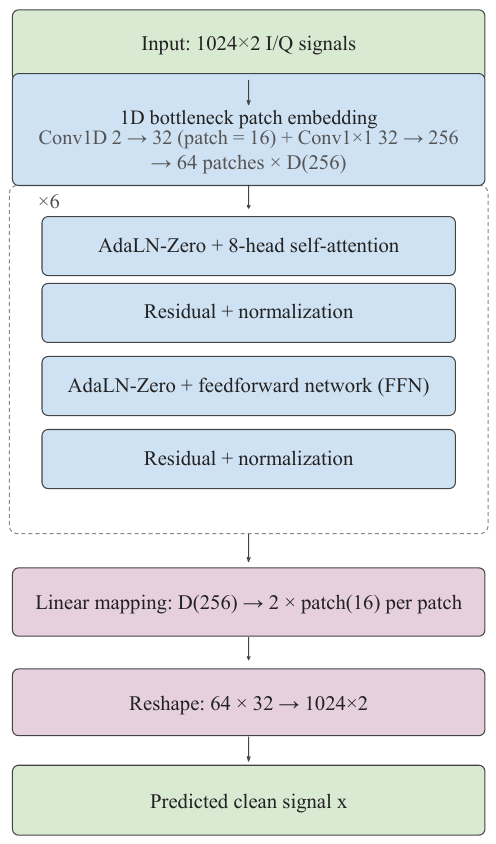}
\caption{Structure of the signal denoising network.}\label{fig:2}
\end{figure}

Through the flow matching training process described above, clean signal examples $x$ are provided as input. The network parameters $\theta$ are iteratively updated to minimize the loss function $L$, yielding a signal denoising network $\mathrm{net}_\theta$ with a low-dimensional manifold prior. The network learns the low-dimensional manifold in which clean signals are embedded, enabling it to project arbitrary noisy inputs back onto the clean manifold.

\begin{table}[!htbp]
\centering\small
\caption{Training algorithm for the denoising network.}\label{tab:algorithm1}
\begin{minipage}{0.96\linewidth}
\hrule\vspace{4pt}
\textbf{Input:} Clean signal data $x$.\\[2pt]
\textbf{Output:} Parameters $\theta$ of the trained denoising network.\\[2pt]
\textbf{Initialize} $\mathrm{net}_\theta$ and $\mathrm{epoch}=1$.\\[2pt]
\textbf{While} $\mathrm{epoch}\leq 200$:\\[2pt]
\quad Sample a time $t$ and Gaussian noise $\varepsilon$.\\[2pt]
\quad Construct $x_t=tx+(1-t)\varepsilon$.\\[2pt]
\quad Compute $v=(x-x_t)/(1-t)$.\\[2pt]
\quad Predict $x_\theta=\mathrm{net}_\theta(x_t,t)$.\\[2pt]
\quad Predict $v_\theta=(x_\theta-x_t)/(1-t)$.\\[2pt]
\quad Compute $L=\mathbb E_{t,x,\varepsilon}\lVert v_\theta-v\rVert^2$.\\[2pt]
\quad Perform backpropagation and update the network parameters $\theta$.\\[2pt]
\quad $\mathrm{epoch}=\mathrm{epoch}+1$.\\[2pt]
\textbf{End}\\[2pt]
\textbf{Return} $\theta$.
\vspace{4pt}\hrule
\end{minipage}
\end{table}

\subsection{Denoising Defense Against Adversarial Examples via JiT Projection}

By definition, an adversarial example is obtained by adding a specific, small perturbation $\delta$ to a clean example $x$, thereby deceiving the network into misclassifying it. In Eq. (5), $\theta$ denotes the parameters of the recognition network C, and $l$ denotes the incorrect target class that the model is intended to output. Minimizing the objective function $\arg\min_\delta\lVert\delta\rVert_2$ aims to make the model output the target class with the smallest possible perturbation. The constraint $x+\delta\in[0,1]^m$ ensures that each input component remains within the allowable range.

\begin{equation}\label{eq:5}
\begin{aligned}&\arg\min_{\delta}\ \lVert\delta\rVert_2\\&\text{subject to}\quad C(x+\delta;\theta)=l,\qquad x+\delta\in[0,1]^m .\end{aligned}
\end{equation}

Regardless of the attack method used, every attack essentially involves solving for a perturbation $\delta$. The computation of this perturbation depends on both the clean example and the target DNN, and the perturbation still exhibits certain noise-like properties. From this perspective, adversarial examples can be regarded as noisy examples, with adversarial perturbations constituting noise obtained by carefully solving optimization problem (5). Any denoising method that brings the denoised result closer to the clean example can therefore provide a degree of adversarial defense. However, blind denoising without prior information is often difficult and inefficient. Identifying a universal prior that is independent of the adversarial attack method is crucial for denoising defense.

From a manifold perspective, if a network learns a mapping that projects any example back onto the low-dimensional manifold of clean data, it can project off-manifold examples onto that manifold, thereby achieving effective denoising-based defense. Moreover, this defense strategy requires no prior information about the methods used to generate adversarial examples and can provide universal defense against various attack types.

Therefore, the training procedure in Section 2.1 yields a network $\mathrm{net}_\theta$ that learns a mapping projecting arbitrary examples back onto the low-dimensional manifold of clean examples. An arbitrary signal example $x_t$ and a denoising start time $t_{\rm start}$ are then fed into this network $\mathrm{net}_\theta$. As time progresses from $t=t_{\rm start}$ to $t=1$, the network gradually projects the input signal example back onto the corresponding point on the clean manifold. This produces a denoised clean example, with the detailed procedure presented in Table 2. This denoising process requires no prior knowledge of the specific attack type and offers effective defense across attack types. Experiments in this study demonstrate that substantial adversarial defense can be achieved through a simple procedure. An arbitrary signal $x_i$ and a denoising start time $t_{\rm start}$ are directly fed into the trained denoising network, and the output denoised signal is classified. This study also presents experimental analysis and discussion of the selection of the denoising start time $t_{\rm start}$.

\begin{table}[!htbp]
\centering\small
\caption{Denoising procedure.}\label{tab:algorithm2}
\begin{minipage}{0.96\linewidth}
\hrule\vspace{4pt}
\textbf{Input:} Signal $x_i$ and start time $t_{\rm start}$.\\[2pt]
\textbf{Network:} Trained $\mathrm{net}_\theta$.\\[2pt]
\textbf{Procedure:} Project the input onto the clean-signal manifold \quad $x_0=\mathrm{net}_\theta(x_i,t_{\rm start})$.\\[2pt]
\textbf{Output:} Denoised clean signal $x_0$.
\vspace{4pt}\hrule
\end{minipage}
\end{table}

\section{Experiments}

\subsection{Experimental Setup}

This study uses the subset of the RML2018.01A dataset with a signal-to-noise ratio (SNR) of 10 dB. The subset covers 24 modulation types: OOK, 4ASK, 8ASK, BPSK, QPSK, 8PSK, 16PSK, 32PSK, 16APSK, 32APSK, 64APSK, 128APSK, 16QAM, 32QAM, 64QAM, 128QAM, 256QAM, AM-SSB-WC, AM-SSB-SC, AM-DSB-WC, AM-DSB-SC, FM, GMSK, and OQPSK. The examples are I/Q signals with a shape of 1024 $\times$ 2.

\begin{figure}[!htbp]
\centering
\includegraphics[height=.48\textheight,keepaspectratio]{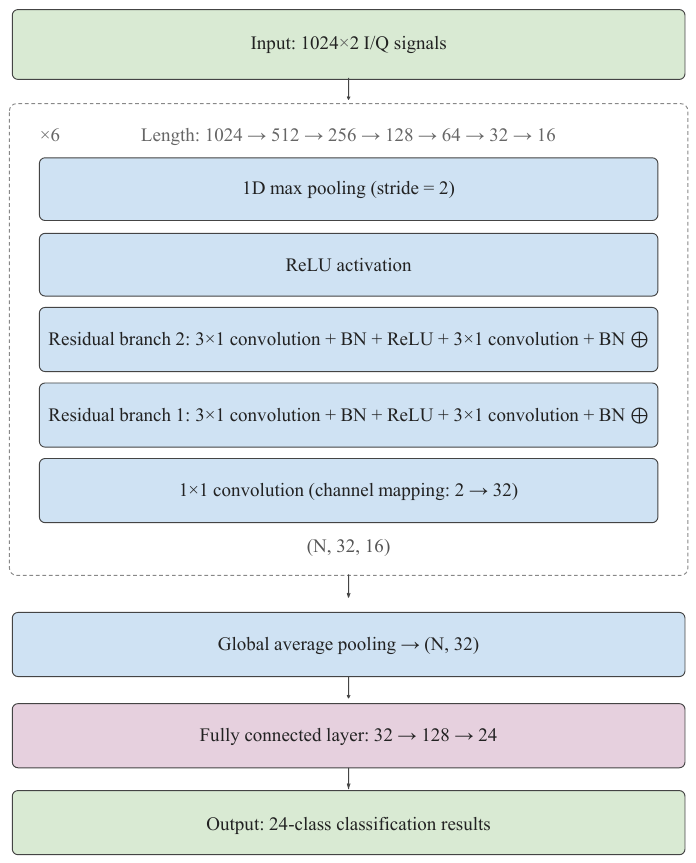}
\caption{Architecture of the ResNet-1D classifier.}\label{fig:3}
\end{figure}

The classifier is a ResNet comprising six residual blocks~\cite{ref25}, with its architecture shown in Fig. 3. To accommodate the one-dimensional nature of signal data, each residual block contains a 1 $\times$ 1 convolution for channel mapping, two residual connections, and max pooling with a stride of 2. Each residual connection contains two 3 $\times$ 1 convolution layers with batch normalization (BN) and ReLU. The network has two input channels, and all hidden layers have 32 channels. Max pooling reduces the sequence length from 1024 to 16. Global pooling is followed by a fully connected layer that produces a 24-class output.

Adversarial training is used as a defense baseline by fine-tuning the original classifier with adversarial examples generated using PGD ($\delta=0.16$, an $L^\infty$ constraint, a step size of $\alpha=\delta/4$, and 10 iterations). The attack methods and strengths used for testing are listed in Table 3. The remaining parameters are set as follows. PGD and BIM~\cite{ref26} use a step size of $\alpha=\delta/4$ and 10 iterations. MIM~\cite{ref27} introduces a momentum term into BIM, with a decay factor of $\mu=1.0$. It accumulates past gradient directions to enhance perturbation transferability. AutoAttack~\cite{ref28} comprises four complementary attack strategies, APGD-CE, APGD-DLR, FAB, and Square, and uses 10 iterations. The overshoot parameter of DeepFool~\cite{ref29} is fixed at 0.0001. For the C\&W attack~\cite{ref30}, the optimizer learning rate is 0.01, the confidence parameter is 0.5, and the initial trade-off constant is 0.01. NES~\cite{ref31} uses a population size of 30 and 50 optimization steps.

This study uses clean accuracy and adversarial robust accuracy to evaluate the performance of defense methods on adversarial examples generated by various attacks. All experiments are conducted on a workstation equipped with an Intel Xeon Silver 4210 processor and an NVIDIA RTX 3090 GPU, running Windows 10. The experimental code is written in Python 3.9.2 and implemented using PyTorch 2.8.0. GPU acceleration is enabled using a PyTorch build with CUDA 12.6 support.

\subsection{Defense Performance at Different Denoising Start Times}

In the JiT model, $t=0$ represents pure noise, whereas $t=1$ represents a clean example. Denoising proceeds from $t=0$ to $t=1$. Adversarial examples already lie near the clean manifold. They can therefore be projected onto the clean manifold from a relatively high denoising start time $t_{\rm start}$, without starting from pure noise. The $t_{\rm start}$ can characterize the noise magnitude in adversarial examples. A larger $t_{\rm start}$ indicates that an adversarial example is closer to a clean example. Experiments show that an excessively high start time $t_{\rm start}$ leads to incomplete removal of adversarial noise, reducing adversarial robust accuracy. If the start time $t_{\rm start}$ is excessively low, excessive removal of adversarial noise damages clean examples and also reduces adversarial robust accuracy. Tests using different attack methods and strengths were conducted on the subset of the RML2018.01A dataset with an SNR of 10 dB. The denoising results at different denoising start times are presented in Table 3.

Table 3 shows that, for gradient-based attacks (PGD/MIM/BIM), clean-example accuracy increases monotonically as $t_{\rm start}$ increases. Adversarial robust accuracy, however, first increases and then decreases. Within a given attack type, a higher attack strength corresponds to a smaller optimal value of $t_{\rm start}$. This reveals the trade-off between denoising strength and signal fidelity in JiT denoising defense. Smaller values of $t_{\rm start}$ provide stronger denoising but damage clean signals. Larger values preserve more details of clean signals but cannot sufficiently remove strong perturbations. MIM presents a notable exception. It generates highly transferable perturbations by accumulating gradient directions through momentum. The distribution of these perturbations poses a greater challenge to denoising defense. Across various attack strengths, defense performance against MIM is consistently lower than that against BIM and PGD at the same strength. For the optimization-based attacks DeepFool and C\&W and the black-box attack NES, defense performance improves monotonically as $t_{\rm start}$ increases. This indicates that the perturbations generated by these three attacks are relatively concentrated and lie close to the clean manifold. Defense performance is best near $t_{\rm start}=0.900$ and still tends to improve as $t_{\rm start}$ increases.

\subsection{Comparison with Adversarial Training}

When the attack type and strength are known, the optimal $t_{\rm start}$ can be selected accordingly to further improve defense performance. However, in practical defense scenarios, attack methods and strengths are often diverse, variable, and unpredictable. Consequently, a fixed denoising start time can only be selected empirically. Considering both clean accuracy and adversarial robust accuracy, this study compares denoising defense at $t_{\rm start}=0.850$ with adversarial training under the same attacks. The comparison shows that the proposed denoising defense method provides relatively stable defense performance as attack strength varies. Its defense advantage becomes more pronounced as attack strength increases.

Fig. 4 shows that, compared with PGD-based adversarial training, the JiT denoising model achieves relative improvements in both adversarial robust accuracy and clean-example classification accuracy. The relative improvement in adversarial robust accuracy is at least 20.58\% under PGD attacks at various strengths. The relative improvement in clean-example classification accuracy is 10.20\%. Fig. 5 clearly shows that adversarial training generalizes poorly to unseen attacks. The adversarial training method used in this study is trained on adversarial examples generated by PGD. Across various attack strengths, the proposed method achieves relative improvements in adversarial robust accuracy of at least 21.80\%, 4.09\%, and 42.91\% against BIM, MIM, and AutoAttack, respectively. For optimization-based attacks, the proposed method exhibits an even more pronounced advantage over adversarial training in its ability to defend against attacks unseen by the network. Under DeepFool attacks at various strengths, it achieves a relative improvement of at least 31.53\% in adversarial robust accuracy. Under the stronger attack setting $\mathrm{steps}=5$, the adversarial robust accuracy of adversarial training drops to 30.16\%, whereas the proposed method maintains an adversarial robust accuracy of 60.12\%. Under C\&W attacks at various strengths, the proposed method achieves a relative improvement of at least 30.07\% in adversarial robust accuracy. Under the stronger attack setting of max iterations=9, the adversarial robust accuracy of adversarial training drops to 32.12\%. The proposed method still maintains an adversarial robust accuracy of 61.25\%. For the NES black-box attack, the proposed method achieves a relative improvement of at least 29.48\% in adversarial robust accuracy over adversarial training across different attack strengths. This confirms its effective defense capability in black-box attack scenarios.

The fundamental reason for these observations is that adversarial training is performed using PGD under an $L^\infty$ constraint. The robustness learned by the network primarily addresses the threat model predefined during training. When the attack method changes to DeepFool, C\&W, or NES, the network's adversarial robustness decreases markedly. These attacks differ from those used in adversarial training in their perturbation constraints, optimization objectives, and methods of obtaining attack information. In sharp contrast, JiT denoising defense does not rely on assumptions about attack types. It directly learns a mapping that projects arbitrary examples back onto the low-dimensional manifold of clean examples. Through manifold projection, it eliminates deviations caused by any type of perturbation, giving JiT denoising an inherent ability to defend across attack types. In adapting to unknown attacks, JiT denoising exhibits markedly better defense performance than adversarial training.

For AMC tasks in complex electromagnetic environments, adversarial defense networks with low training costs are easier to deploy and can reduce reliance on high-performance hardware. This study measures training cost in GPU hours (GPU$\cdot$h). Table 4 compares the training costs of JiT denoising network training and adversarial training.

\begingroup
\fontsize{8.2}{9.6}\selectfont
\setlength{\tabcolsep}{3.8pt}
\renewcommand{\arraystretch}{1.10}
\setlength{\LTcapwidth}{\linewidth}
\makeatletter
\renewcommand{\fnum@table}{\small\tablename~\thetable}
\makeatother
\begin{longtable}{@{}>{\raggedright\arraybackslash}p{0.95in}>{\centering\arraybackslash}p{0.95in}*{9}{>{\centering\arraybackslash}p{0.37in}}@{}}
\caption{Adversarial robust accuracy (\%) of denoising defense at different denoising start times.}\label{tab:accuracy}\\
\toprule
Attack & Strength & 0.700 & 0.725 & 0.750 & 0.775 & 0.800 & 0.825 & 0.850 & 0.875 & 0.900\\
\midrule
\endfirsthead
\multicolumn{11}{l}{\small Table \thetable\ (continued)}\\
\toprule
Attack & Strength & 0.700 & 0.725 & 0.750 & 0.775 & 0.800 & 0.825 & 0.850 & 0.875 & 0.900\\
\midrule
\endhead
\midrule\multicolumn{11}{r}{Continued on next page}\\
\endfoot
\bottomrule
\endlastfoot
Clean & -- & 37.69 & 40.39 & 43.62 & 48.76 & 53.50 & 61.93 & 70.00 & 76.53 & 80.39\\
PGD & $\varepsilon$=0.100 & 38.97 & 40.77 & 42.93 & 43.50 & 44.97 & 44.30 & 42.30 & 40.80 & 37.89\\
 & $\varepsilon$=0.125 & 38.10 & 40.37 & 42.50 & 43.10 & 43.97 & 43.07 & 40.80 & 40.10 & 37.20\\
 & $\varepsilon$=0.150 & 37.47 & 40.17 & 41.57 & 42.03 & 41.33 & 41.10 & 39.83 & 39.40 & 37.07\\
 & $\varepsilon$=0.175 & 37.40 & 39.90 & 40.13 & 40.93 & 39.77 & 38.93 & 38.87 & 38.60 & 35.87\\
 & $\varepsilon$=0.200 & 36.70 & 38.36 & 39.80 & 39.50 & 38.63 & 37.63 & 37.10 & 36.13 & 34.07\\
BIM & $\varepsilon$=0.100 & 40.03 & 41.57 & 43.20 & 44.33 & 44.67 & 43.93 & 40.27 & 37.30 & 34.47\\
 & $\varepsilon$=0.125 & 39.43 & 41.30 & 42.97 & 43.37 & 43.73 & 41.24 & 38.57 & 35.27 & 32.93\\
 & $\varepsilon$=0.150 & 38.97 & 40.17 & 41.77 & 41.30 & 40.30 & 39.17 & 37.27 & 35.05 & 29.00\\
 & $\varepsilon$=0.175 & 38.30 & 40.10 & 40.50 & 40.03 & 39.20 & 38.87 & 36.37 & 34.80 & 27.87\\
 & $\varepsilon$=0.200 & 37.47 & 39.83 & 39.23 & 39.00 & 38.73 & 37.63 & 35.40 & 33.70 & 24.70\\
MIM & $\varepsilon$=0.100 & 39.40 & 40.80 & 41.67 & 40.53 & 39.23 & 37.60 & 34.60 & 32.88 & 29.50\\
 & $\varepsilon$=0.125 & 39.23 & 39.73 & 39.93 & 38.07 & 37.27 & 36.13 & 33.37 & 31.47 & 27.43\\
 & $\varepsilon$=0.150 & 38.33 & 38.67 & 37.50 & 37.10 & 36.87 & 35.77 & 32.40 & 30.63 & 22.97\\
 & $\varepsilon$=0.175 & 36.17 & 36.87 & 36.43 & 35.27 & 34.27 & 33.20 & 31.00 & 27.23 & 20.43\\
 & $\varepsilon$=0.200 & 28.83 & 28.87 & 26.90 & 26.40 & 25.77 & 24.30 & 22.83 & 21.67 & 19.17\\
AutoAttack & $\varepsilon$=0.100 & 32.33 & 34.37 & 36.27 & 40.27 & 41.00 & 44.77 & 46.73 & 48.57 & 49.30\\
 & $\varepsilon$=0.125 & 32.00 & 33.97 & 36.03 & 40.10 & 40.40 & 44.27 & 45.23 & 48.23 & 48.57\\
 & $\varepsilon$=0.150 & 31.70 & 33.20 & 35.90 & 39.80 & 40.23 & 42.60 & 43.40 & 47.97 & 47.47\\
 & $\varepsilon$=0.175 & 29.23 & 30.10 & 32.00 & 34.67 & 36.28 & 38.40 & 40.27 & 44.27 & 43.47\\
 & $\varepsilon$=0.200 & 27.17 & 28.43 & 30.23 & 32.90 & 34.67 & 35.57 & 37.83 & 39.20 & 37.43\\
DeepFool & steps=1 & 37.85 & 39.25 & 43.56 & 48.98 & 52.85 & 59.72 & 62.20 & 67.98 & 73.95\\
 & steps=2 & 36.75 & 39.00 & 43.04 & 47.90 & 51.75 & 57.58 & 61.92 & 67.68 & 73.71\\
 & steps=3 & 35.92 & 38.58 & 42.83 & 47.71 & 51.46 & 56.00 & 61.12 & 67.38 & 72.50\\
 & steps=4 & 34.50 & 37.75 & 41.79 & 46.88 & 50.62 & 55.17 & 60.71 & 66.75 & 72.08\\
 & steps=5 & 34.22 & 37.58 & 41.42 & 46.50 & 50.46 & 55.00 & 60.12 & 66.58 & 71.50\\
C\&W & iter.=1 & 37.90 & 40.63 & 42.17 & 48.83 & 50.03 & 56.77 & 63.67 & 67.47 & 72.37\\
 & iter.=3 & 37.83 & 40.50 & 42.03 & 47.77 & 49.17 & 55.87 & 63.00 & 66.77 & 71.77\\
 & iter.=5 & 36.80 & 39.73 & 41.17 & 47.00 & 48.47 & 54.00 & 62.83 & 66.03 & 71.13\\
 & iter.=7 & 36.30 & 39.47 & 40.57 & 46.13 & 47.67 & 53.50 & 61.77 & 65.70 & 70.97\\
 & iter.=9 & 36.33 & 39.00 & 40.25 & 45.17 & 46.67 & 53.02 & 61.25 & 65.22 & 70.67\\
NES & $\varepsilon$=5 & 37.97 & 40.03 & 42.93 & 46.43 & 51.70 & 59.23 & 66.23 & 72.43 & 78.20\\
 & $\varepsilon$=10 & 37.63 & 39.67 & 42.56 & 45.18 & 50.87 & 57.81 & 61.25 & 67.83 & 69.08\\
 & $\varepsilon$=15 & 36.87 & 38.43 & 41.53 & 44.44 & 47.45 & 49.33 & 52.30 & 55.43 & 57.20\\
 & $\varepsilon$=20 & 36.17 & 37.83 & 40.50 & 42.17 & 44.23 & 45.83 & 46.42 & 48.83 & 52.08\\
 & $\varepsilon$=25 & 35.25 & 37.17 & 38.42 & 39.67 & 41.92 & 42.75 & 43.00 & 47.42 & 45.83\\
\end{longtable}
\endgroup

\begin{figure}[!htbp]
\centering
\includegraphics[width=.82\linewidth,height=.34\textheight,keepaspectratio]{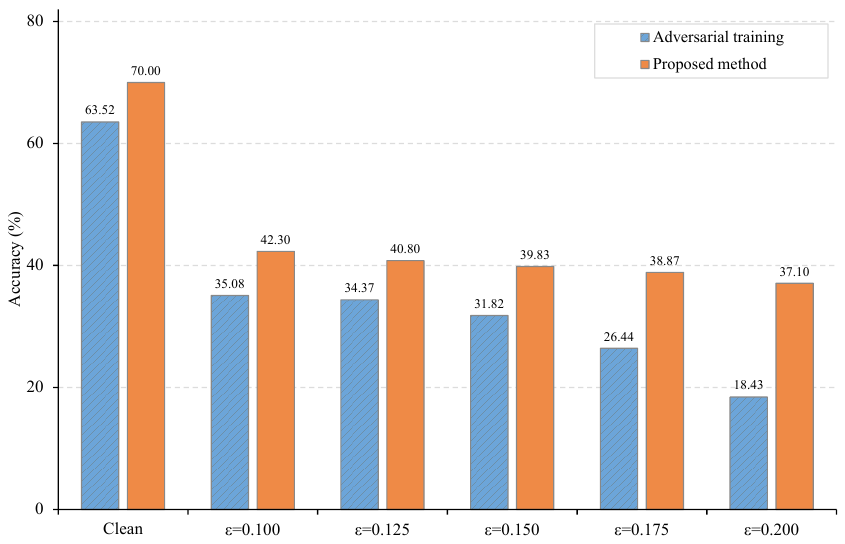}
\caption{Comparison of defense accuracy (\%) between JiT denoising ($t_{\rm start}=0.850$) and adversarial training under PGD attacks.}\label{fig:4}
\end{figure}

\begin{figure}[!htbp]
\centering
\includegraphics[width=.82\linewidth,height=.45\textheight,keepaspectratio]{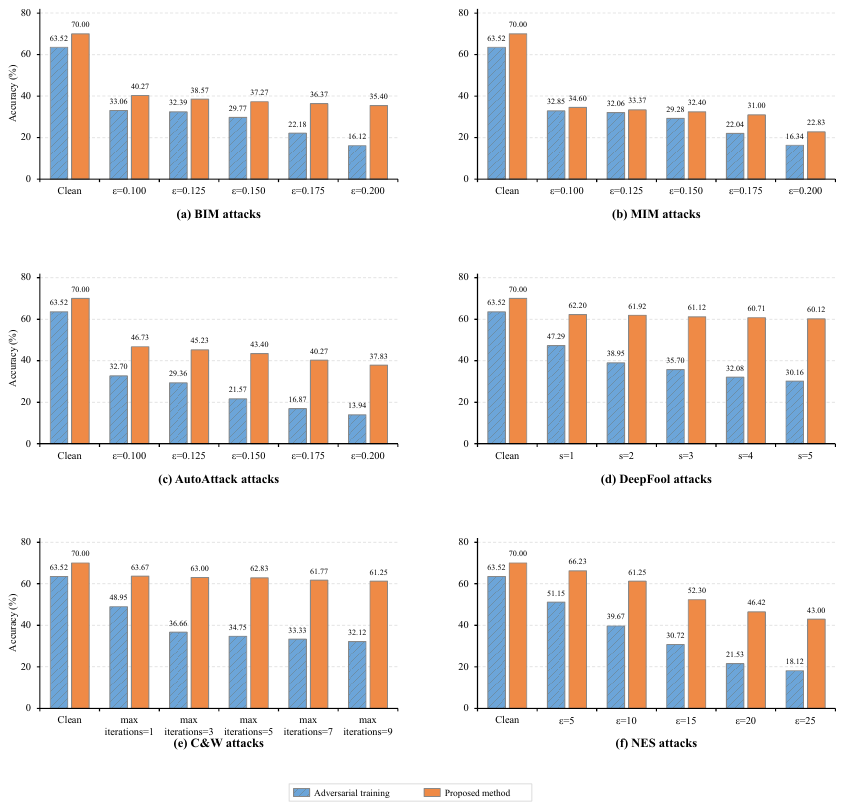}
\caption{Comparison of defense accuracy (\%) between JiT denoising ($t_{\rm start}=0.850$) and adversarial training under other attacks.}\label{fig:5}
\end{figure}

As shown in Table 4, the slow generation of adversarial examples results in a total training cost of 14.68 GPU$\cdot$h for adversarial training. In comparison, the JiT denoising network is trained on clean examples. Its total training time is only 1.21 GPU$\cdot$h, approximately 8.24\% of that required for adversarial training.

\begin{table}[!htbp]

\centering\small

\caption{Comparison of training costs for JiT denoising network training and adversarial training.}\label{tab:cost}

\begin{tabular}{>{\centering\arraybackslash}p{2.8in}>{\centering\arraybackslash}p{1.65in}>{\centering\arraybackslash}p{0.75in}}\toprule

Training method & Total training epochs & GPU$\cdot$h\\\midrule

JiT denoising network training & 200 & 1.21\\

Adversarial training & 200 & 14.68\\

\bottomrule\end{tabular}

\end{table}

\section{Conclusion}

To address adversarial threats to automatic modulation classification of radio signals, this paper proposes an effective denoising defense method against adversarial examples. The proposed method is motivated by the assumption that adversarial examples are perturbed signals lying off the low-dimensional manifold of clean signals. By learning a mapping that projects arbitrary examples back onto the low-dimensional manifold of clean examples, the network enables prior-guided projection denoising onto this manifold. A 1D Transformer denoising network tailored to signal data is designed according to the characteristics of one-dimensional radio signal I/Q data. It is trained on clean signal examples using flow matching. The resulting denoising network learns a mapping that projects arbitrary examples back onto the low-dimensional manifold of clean-example data. On this basis, arbitrary examples are fed into the trained network and directly projected onto the low-dimensional manifold of clean examples for denoising. Experiments on the RML2018.01A dataset evaluate the proposed defense method using a single empirically selected, fixed denoising start time. Compared with PGD-based adversarial training, the method achieves a relative improvement of at least 20.58\% in adversarial robust accuracy under PGD attacks at various strengths. It also achieves a relative improvement of 10.20\% in clean-example recognition accuracy. For attack types unseen during adversarial training, the proposed method exhibits more pronounced defense advantages. In particular, at higher attack strengths, it achieves a relative improvement of at least 39.72\% in adversarial robust accuracy. The method does not depend on any specific attack method and performs ``direct denoising'' using adversarial examples as denoising starting points. It demonstrates the advantages of low training costs, straightforward implementation, and strong defense capabilities across attack types.

\end{document}